\documentclass{article}

\usepackage[main, final]{neurips_2026}
\usepackage[utf8]{inputenc}
\usepackage[T1]{fontenc}
\usepackage[hidelinks,backref=page]{hyperref}
\usepackage{diagbox}
\usepackage{annotate-equations}
\usepackage{booktabs}
\usepackage{array}
\usepackage{makecell}
\usepackage{graphicx}
\usepackage{adjustbox}
\usepackage{url}
\usepackage[table]{xcolor}
\usepackage{amsfonts}
\usepackage{nicefrac}
\usepackage{amsmath}
\usepackage{accents}
\usepackage{xcolor}
\usepackage{setspace}
\usepackage{bbm}
\usepackage{colortbl}
\usepackage{microtype}
\usepackage{graphicx}
\usepackage{subfigure}
\usepackage{booktabs}
\usepackage{amssymb}
\usepackage[most]{tcolorbox}
\usepackage{mathtools}
\usepackage{amsthm}
\usepackage[capitalize,noabbrev]{cleveref}

\definecolor{darkred}{RGB}{150,0,0}
\definecolor{darkgreen}{RGB}{0,150,0}
\definecolor{darkblue}{RGB}{0,0,150}
\hypersetup{colorlinks=true, linkcolor=darkred, citecolor=darkblue, urlcolor=darkblue}

\makeatletter

\theoremstyle{plain}

\theoremstyle{definition}

\theoremstyle{remark}

\makeatletter
\makeatletter
\DeclareRobustCommand\onedot{\futurelet\@let@token\@onedot}
\def\@onedot{\ifx\@let@token.\else.\null\fi\xspace}

\makeatother

\makeatletter

\usepackage{etoolbox}
\usepackage{xspace}
\usepackage{enumitem}
\usepackage{setspace}
\usepackage{nccmath}
\usepackage{bm}
\usepackage{multirow}
\usepackage{afterpage}
\usepackage{babel}
\usepackage{algpseudocode}
\usepackage{subcaption}

\title{Shaping Useful Noise: Energy Distributions\\Predict Visual Pretraining Quality}

\author{
  Ching Lam Choi$^{1}$
  \quad
  Antonio Torralba$^{1}$
  \quad
  Phillip Isola$^{1}$ 
  \quad
  Stefanie Jegelka$^{1,2}$\\
  $^1$CSAIL, Department of EECS, Massachusetts Institute of Technology\\
  $^2$School of CIT, MCML, MDSI, Technical University of Munich\\
  \texttt{\{\href{mailto:chinglam@mit.edu}{\textcolor{black}{chinglam}}, \href{mailto:torralba@mit.edu}{\textcolor{black}{torralba}}, \href{mailto:phillipi@mit.edu}{\textcolor{black}{phillipi}}\}@mit.edu},\quad \texttt{stefanie.jegelka@tum.de} 
}

\begin{document}

\maketitle

\begin{abstract}
Synthetic data can scale visual pretraining, but current selection criteria only partially explain which datasets train strong representations. We propose using the \emph{energy profile} of a dataset---the distribution of log-probability levels it occupies under a normalised real-image density model---as an audit and curation principle for synthetic data. To operationalise this view, we introduce \emph{Energy-Level Distribution Distance (\textsc{Eld})}, which compares real and synthetic energy distributions. Real images span sparse high-likelihood scenes and dense low-likelihood textures, so useful synthetic data should cover this energy range rather than collapse to likely but low-variation samples. Across diverse synthetic datasets, \textsc{Eld} exposes mismatches missed by FID and predicts pretraining performance beyond FID, LPIPS variation, and feature-space recall, including across CNN and ViT self-supervised settings. We then use \textsc{Eld} to resample and mix synthetic sources, matching the real-image energy histogram while preserving diversity and improving downstream transfer.

\textbf{Project page:}\quad \href{https://chinglamchoi.github.io/papers/noiseAudits}{chinglamchoi.github.io/papers/noiseAudits}\\\textbf{Code:}\quad\href{https://github.com/chinglamchoi/shaping-useful-noise}{github.com/chinglamchoi/shaping-useful-noise}
\end{abstract}

\section{Introduction}
Synthetic visual pretraining turns data choice into a distributional design problem: models can be trained on simulations, procedural images, generator samples, or mixtures thereof, each inducing different visual statistics. Yet current selection criteria often reduce this choice to sample realism or feature-space similarity. These criteria are useful, but they compress datasets into distances, neighbourhoods or low-order feature summaries, and do not ask where samples lie in the probability landscape of real images. This missing axis matters because natural images do not occupy a single typical likelihood level. Normalised density models reveal a broad, content-dependent energy distribution, where sparse simple scenes and dense textured scenes occupy different likelihood regimes \citep{guth2026learning,kadkhodaie2024generalization}. This echoes classical natural-image statistics, which show that real images have structured spectral, wavelet and sparsity properties far from white noise \citep{burton1987color,field1987relations,ruderman1997origins,simoncelli20054,portilla2000parametric}. A synthetic dataset may therefore match real feature moments while exposing a learner to the wrong likelihood regimes; selecting only the most likely samples may likewise remove complex variation that real data contains. We ask whether synthetic data should be selected and designed to match the \emph{distribution of likelihood levels} occupied by real images.

\subsection{Our Contributions}
We make four contributions around a common claim: beyond visual realism or feature-space similarity, \emph{the energy shape of synthetic data is a useful predictor and design principle for visual pretraining.} 

First, we formalise this design axis through Energy-Level Distribution Distance (\textsc{Eld}), together with a percentile variant, which compares real and synthetic datasets through their normalised real-image log-probability distributions. Unlike mean likelihood, \textsc{Eld} measures the full energy profile of a dataset. It penalises synthetic data that drifts into low-likelihood, noise-like regions, as well as data that collapses towards high-likelihood but low-variation samples. This makes \textsc{Eld} complementary to feature-space metrics such as FID, LPIPS variation and recall, which measure moments, perceptual distances or coverage rather than probability-level structure.

Second, we use \textsc{Eld} to audit a broad suite of procedural, statistical, feature-visualisation and generator-based synthetic datasets. The audit reveals three recurring regimes: noise-like datasets that sit in low-likelihood regions, over-concentrated datasets that collapse around high-likelihood but low-variation samples, and broad-but-shifted datasets that span many energy levels while missing the real profile. These regimes expose why feature-space metrics can be incomplete: two datasets may have similar FID, LPIPS variation or recall while presenting very different likelihood structure to the learner. Third, we show that this structure is predictive, not merely descriptive: \textsc{Eld} explains pretraining utility beyond standard metrics, including across CNN and ViT self-supervised settings. Finally, we turn the audit into an intervention. By resampling within synthetic pools and mixing across generators, we match the real-image energy histogram without sacrificing feature diversity, yielding synthetic training sets with stronger downstream transfer.

\begin{figure}[t]
\centering
\includegraphics[width=\textwidth]{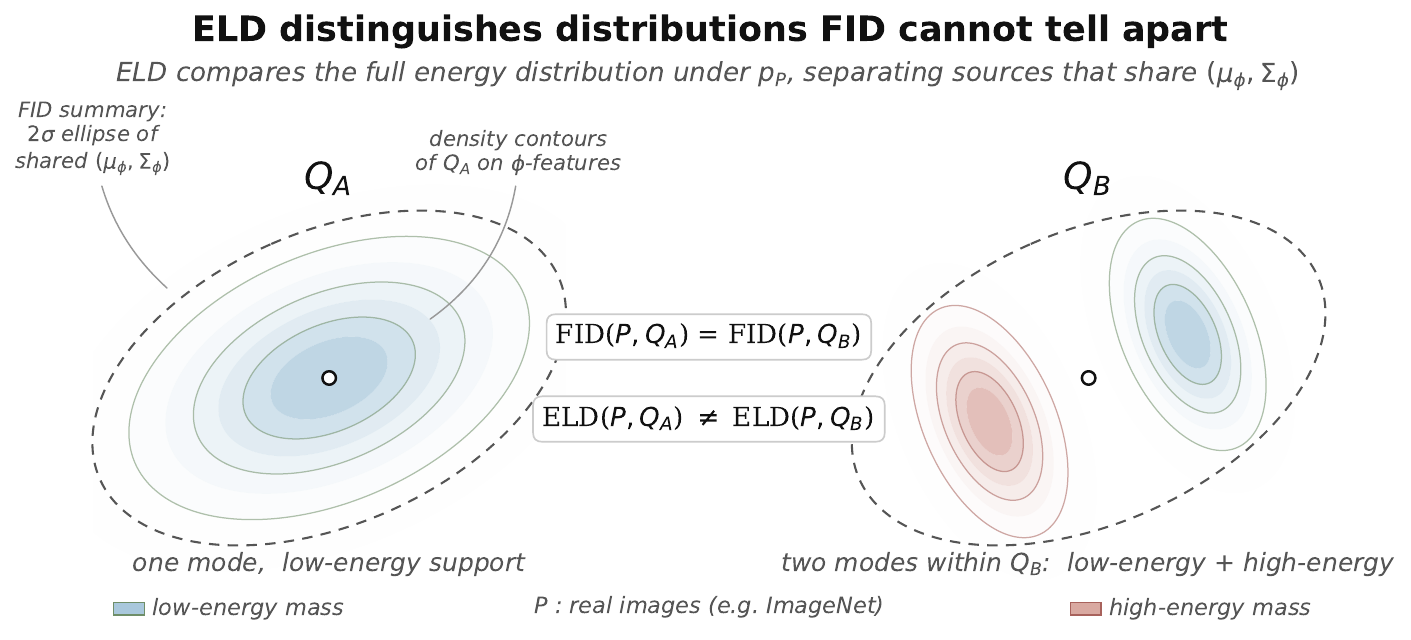}
\caption{\textbf{ELD distinguishes synthetic pretraining datasets FID cannot.} Self-supervised pretraining benefits when synthetic data spans the full likelihood spectrum of real images; sources with identical FID can carry very different energy distributions, which ELD detects.}
\vspace{-10pt}
\label{fig:teaser}
\end{figure}

\subsection{Related Work}

\paragraph{Synthetic data for visual pretraining.}
Synthetic visual data has been explored through several generation mechanisms: simulation and domain randomisation for transfer to real environments \citep{tobin2017domain}, fractal and procedural image datasets for pretraining without photographs \citep{kataoka2020pre}, structured noise and statistical image models for contrastive representation learning \citep{baradad2021learning}, and text-to-image generators for large-scale synthetic representation learning \citep{NEURIPS2023_971f1e59}. These works establish a rich design space for synthetic pretraining data. Our focus is not to introduce a new generator, but to identify a distribution-level property that predicts which generated datasets are useful across modern self-supervised objectives.

\paragraph{Distribution metrics for generated data.}
Most  metrics compare real and generated distributions in a learned feature space. FID fits Gaussian approximations to Inception features \citep{NIPS2017_8a1d6947}; KID uses a kernel distance \citep{binkowski2018demystifying}; precision and recall estimate fidelity and coverage \citep{kynkaanniemi2019improved,naeem2020reliable}; and LPIPS measures perceptual similarity with deep features \citep{zhang2018unreasonable}. These metrics capture fidelity and diversity, but recent work shows their sensitivity to feature choice and Gaussian assumptions \citep{Jayasumana_2024_CVPR}. \textsc{Eld} is orthogonal to these criteria: it compares probability-level structure under a real-image density model rather than distances or moments in representation space.

\paragraph{Image likelihoods and energy structure.}
Energy-based models describe data by negative log density, but calibrated image likelihoods are hard to estimate \citep{hinton1986learning,lecun2006tutorial,song2021train}. Normalising flows give exact likelihoods through invertible transformations \citep{rezende2015variational,dinh2014nice,dinh2017density}, while diffusion and score-based models estimate image distributions through denoising dynamics \citep{sohl2015deep,NEURIPS2020_4c5bcfec,song2021scorebased}. Recent normalised density estimators enable direct analysis of natural-image likelihoods; \citet{guth2026learning} show that real-image log probabilities are broad, content-dependent, and approximately Gumbel-like. We use this likelihood-level view as a diagnostic and curation principle: it identifies which synthetic datasets are mismatched and how to reshape them.

\section{Energy Structure of Natural and Synthetic Images}\label{sec:energy-structure}
A useful synthetic pretraining distribution should reproduce more than the visual appearance of real images. We focus on one structural axis: the distribution of likelihood levels assigned to a dataset by a normalised real-image density model. This energy distribution is rich in natural images; it is not determined by FID or other moment-based feature-space metrics, and separates synthetic sources into three regimes with distinct measurable signatures.

\subsection{The Energy Distribution of Natural Images}
\label{sec:natural-energy}
Let $P$ be a real-image distribution and $p_P$ a normalised density model fitted to $P$. The \emph{real-image energy} $E_P(x)=-\log p_P(x)$ scores how typical $x$ is under the model: low for images $p_P$ assigns high likelihood, high for images $p_P$ assigns low likelihood. The pushforward $(E_P)_{\#}P$ records how often real images occur at each likelihood level. This energy distribution is broad: \citet{guth2026learning} show that on ImageNet it is approximately Gumbel-like, with sparse simple scenes at the low-energy/high-likelihood end and dense textured scenes forming a high-energy/low-likelihood tail, consistent with the statistical inhomogeneity of natural images \citep{field1987relations,ruderman1997origins,simoncelli20054,portilla2000parametric}.

\begin{figure}[t]
    \centering
     \includegraphics[width=\textwidth]{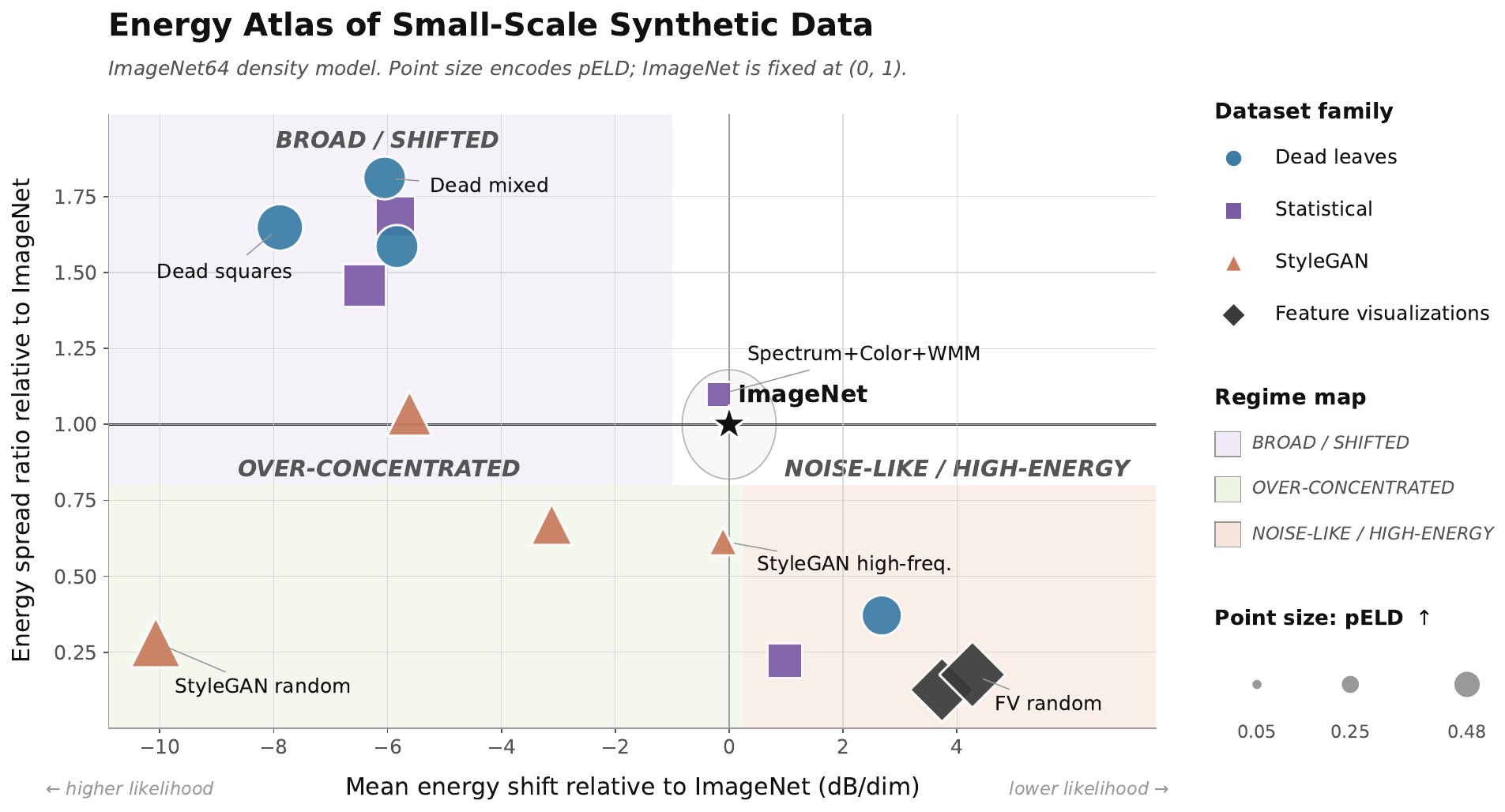}
    \caption{\textbf{Small-scale energy audit.} Synthetic datasets scored under a grayscale ImageNet64 density model. Axes show mean energy shift and spread relative to ImageNet, while point size denotes \textsc{pEld}. The atlas provides a coarse diagnostic view of the regimes expanded in Figure~\ref{fig:eld-method}: noise-like sources are shifted toward high-energy regions, over-concentrated sources occupy too narrow a likelihood band, and broad/shifted sources cover a wider range but remain displaced from the real energy profile.}
    \vspace{-10pt}
    \label{fig:energy-regimes}
\end{figure}

The implication for synthetic pretraining is concrete. A useful synthetic distribution should expose the learner to the same range of likelihood levels as real data, not merely maximise average likelihood. Pushing $\mathbb{E}_Q[\log p_P]$ upward can collapse $Q$ to the mode of $p_P$; ignoring likelihood can place mass arbitrarily far in the high-energy tail. Either failure appears as a mismatch in $(E_P)_{\#}Q$. The relevant comparison is therefore not a scalar summary such as mean likelihood, but the full energy profiles $(E_P)_{\#}P$ and $(E_P)_{\#}Q$, which Section~\ref{sec:eld-definition} measures.

\begin{figure}[t]
    \centering
    \includegraphics[width=1.05\textwidth]{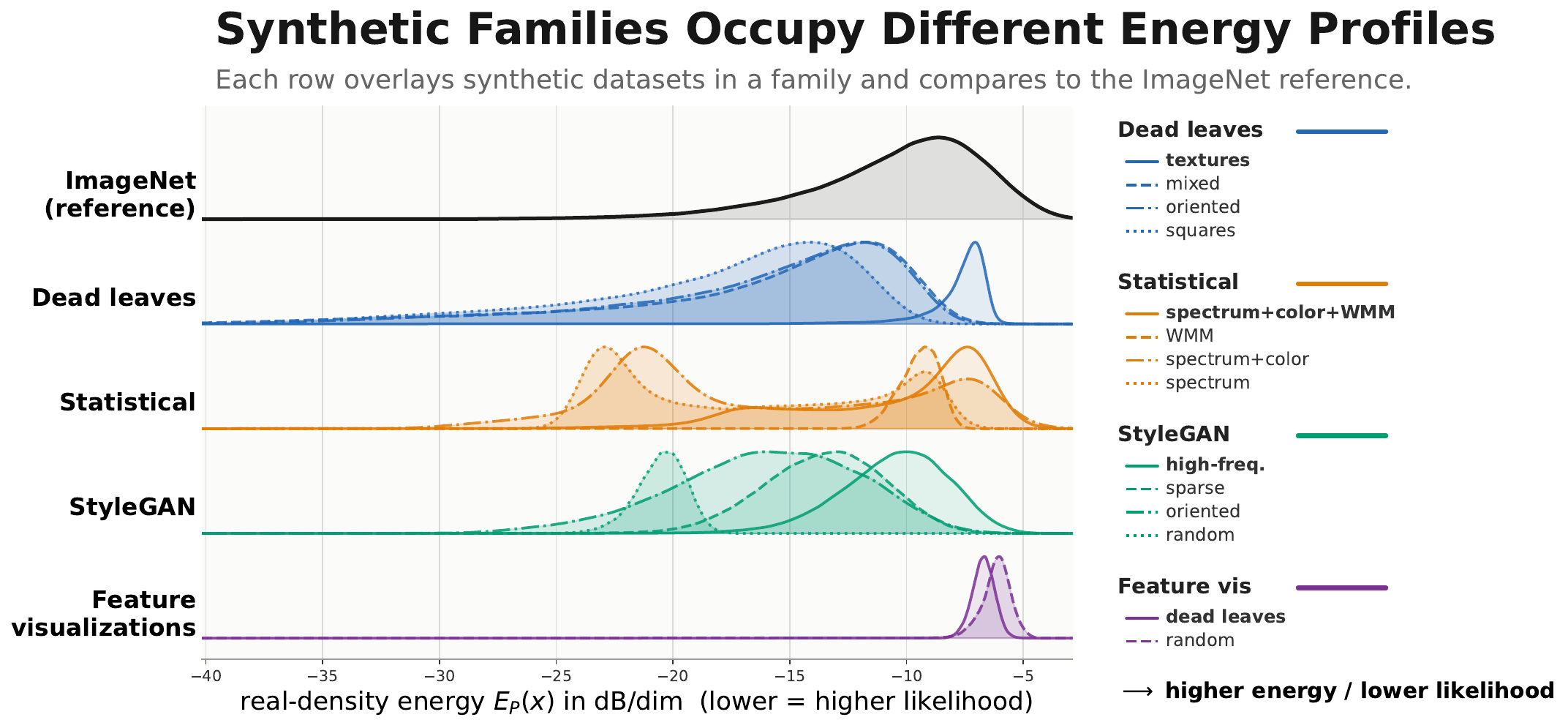}
    \caption{\textbf{Energy mismatch is structured.} Synthetic sources differ in how they depart from ImageNet: some collapse into narrow high-likelihood bands, others extend into low-likelihood tails, and others remain broad but displaced. Effective synthetic pretraining may require matching the full probability-level profile of natural images, not simply maximising likelihood or feature realism.}
    \vspace{-10pt}
    \label{fig:eld-method}
\end{figure}

\subsection{Feature Moments Do Not Determine Energy Profiles}\label{sec:fid-blind}
FID compares the means and covariances of a fixed feature map. These moments need not determine how samples occupy energy levels, as in Figure~\ref{fig:teaser}. For example, under the identity feature map, let
\[
 Q_1=\tfrac12\delta_{-1}+\tfrac12\delta_1,
 \qquad
 Q_2=\tfrac14\delta_{-\sqrt{2}}+\tfrac12\delta_0
       +\tfrac14\delta_{\sqrt{2}}.
\]
Both have mean zero and variance one, and therefore identical FID to any fixed reference. Under the normalised density $p(x)=Z^{-1}\exp(-x^4)$, however, $Q_1$ has energy $1+\log Z$ almost surely, whereas $Q_2$ has energies $\log Z$ and $4+\log Z$ with equal probability. Thus identical feature moments can conceal different energy profiles. Appendix~\ref{app:fid_fibres} gives the general statement.

\subsection{A Diagnostic Taxonomy of Synthetic Energy Mismatch}
\label{sec:synthetic-regimes}
If FID does not determine energy matching, the remaining question is empirical: do synthetic datasets differ along this axis? Figure~\ref{fig:energy-regimes} visualises representative sources scored under a shared real-image density model $p_P$. Let $\mu_P,\sigma_P$ denote the mean and standard deviation of $E_P$ on real data, and $\mu_Q,\sigma_Q$ the corresponding quantities for a synthetic source. Three regimes recur. \emph{Noise-like} datasets have $\mu_Q-\mu_P$ large and positive relative to $\sigma_P$, placing mass in the high-energy tail: the real model rates these images as implausible. \emph{Over-concentrated} datasets have small $\sigma_Q/\sigma_P$, often with $\mu_Q\le\mu_P$: samples may be individually plausible, but occupy too narrow a slice of natural-image structure. \emph{Broad/shifted} datasets have substantial mismatch in either spread or location, with $|\mu_Q-\mu_P|$ and/or $\sigma_Q/\sigma_P$ far from the real reference. Mixed cases interpolate between these regimes. Figure~\ref{fig:eld-method} shows the same regimes at the level of full energy distributions, making the shift, compression, and tail mismatches visible beyond the two summary statistics.

These diagnostics depend on the calibration of $p_P$; Section~\ref{sec:eld-definition} introduces a percentile-rank variant that removes this dependence. As shown in Section~\ref{sec:fid-blind}, datasets in different regimes can have similar FID despite different energy arrangements, and Section~\ref{sec:experiments} confirms this empirically. These regimes show why ``realism'' is incomplete for synthetic pretraining: a dataset can be too unlikely, too concentrated near high-likelihood regions, or mismatched across likelihood levels. The target is the full real-image energy distribution, not a scalar likelihood or feature-moment score.

\section{\textsc{ELD}: Energy-Level Distribution Distance}
Section~\ref{sec:fid-blind} shows that feature-moment metrics do not determine a dataset's energy profile. We now define a metric that compares datasets along this axis directly.

\subsection{Definition and Estimation}
\label{sec:eld-definition}

Given a normalised real-image density model $p_P$ with energy $E_P(x)=-\log p_P(x)$, the \emph{Energy-Level Distribution Distance} between real and synthetic distributions $P,Q$ is
\begin{equation}
    \mathrm{ELD}(P,Q)\;=\;W_1\!\bigl((E_P)_{\#}P,\,(E_P)_{\#}Q\bigr),
\end{equation}
the Wasserstein-1 distance between their one-dimensional energy distributions. In one dimension, $W_1$ equals the integral distance between empirical CDFs \citep{villani2021topics}, making \textsc{ELD} simple to estimate from sorted energy scores. We use $W_1$ rather than KL because it remains well-defined under support mismatch and returns a distance in the units of the energy score.

Raw energies depend on the calibration of $p_P$, so we also use a percentile version. Let $F_P$ denote the CDF of real-image energies $E_P(x)$ for $x\sim P$. We replace each energy by its real-energy rank $F_P(E_P(x))\in[0,1]$. By construction, $(F_P\circ E_P)_{\#}P=\mathrm{Unif}(0,1)$, so
\begin{equation}
    \mathrm{pELD}(P,Q)\;=\;W_1\!\bigl(\mathrm{Unif}(0,1),\,(F_P\circ E_P)_{\#}Q\bigr).
\end{equation}
For a continuous real-energy CDF $F_P$, the probability integral transform yields uniform real-energy percentiles $u(x)=F_P(E_P(x))$. In finite samples, we estimate $F_P$ from a real calibration set; independent real samples are therefore only approximately uniform. \textsc{pEld} is bounded in $[0,1/2]$ and invariant to strictly increasing recalibrations of the energy score applied consistently to real and synthetic samples. At the population level, if $F_P$ is strictly increasing on the combined real and synthetic energy support, it vanishes exactly when their energy distributions coincide. This removes score-scale dependence, but not changes in image ordering induced by a different scorer. Unless otherwise stated, we report \textsc{pEld} as \textsc{Eld}. Raw \textsc{Eld} is expressed in the units of its energy score; our plotted energies use dB per dimension.

We instantiate $p_P$ with the normalised ImageNet density estimator of \citet{guth2026learning}, trained using a diffusion-inspired dual-score-matching objective. It directly provides tractable image log-density estimates, without requiring sampling from the scorer. Once energies are computed, \textsc{Eld} requires only empirical CDFs; Section~\ref{sec:experiments} shows that a few thousand samples per distribution suffice for stable rankings and evaluates robustness to the choice of density scorer. The main cost is energy evaluation, amortised across all synthetic datasets scored under the same $p_P$.

\subsection{Interpreting \textsc{Eld}}
\label{sec:eld-captures}
\textsc{Eld} turns the qualitative regimes of Section~\ref{sec:synthetic-regimes} into a single distributional discrepancy. If a synthetic dataset is shifted into the high-energy tail, compressed around high-likelihood modes, or broad but displaced from the real profile, its empirical energy CDF departs from that of real images. The Wasserstein distance integrates these departures across the full likelihood axis, so the metric penalises datasets that are too unlikely, too concentrated, or likely at the wrong frequencies. This is the distinction missed by mean likelihood: \textsc{Eld} does not reward making every sample as likely as possible, but rather matching how real images populate likelihood levels.

This interpretation also clarifies the role of \textsc{Eld}. Rather than replacing semantic or feature-space coverage metrics, \textsc{Eld} isolates a complementary question: whether a dataset populates the same likelihood levels as real images under a shared density model. Two datasets may match in energy profile while differing in semantics or feature diversity; conversely, they may look similar to feature-moment metrics while occupying very different regions of the real-image likelihood axis. We therefore report \textsc{Eld} alongside FID, LPIPS variation, and recall in Section~\ref{sec:experiments}. Figure~\ref{fig:eld-method} illustrates the measured object: when scored by the same ImageNet density model, synthetic families induce distinct energy profiles with different shifts, spreads, and tail structure relative to ImageNet.

\begin{figure}[t!]
\centering
\includegraphics[width=\linewidth]{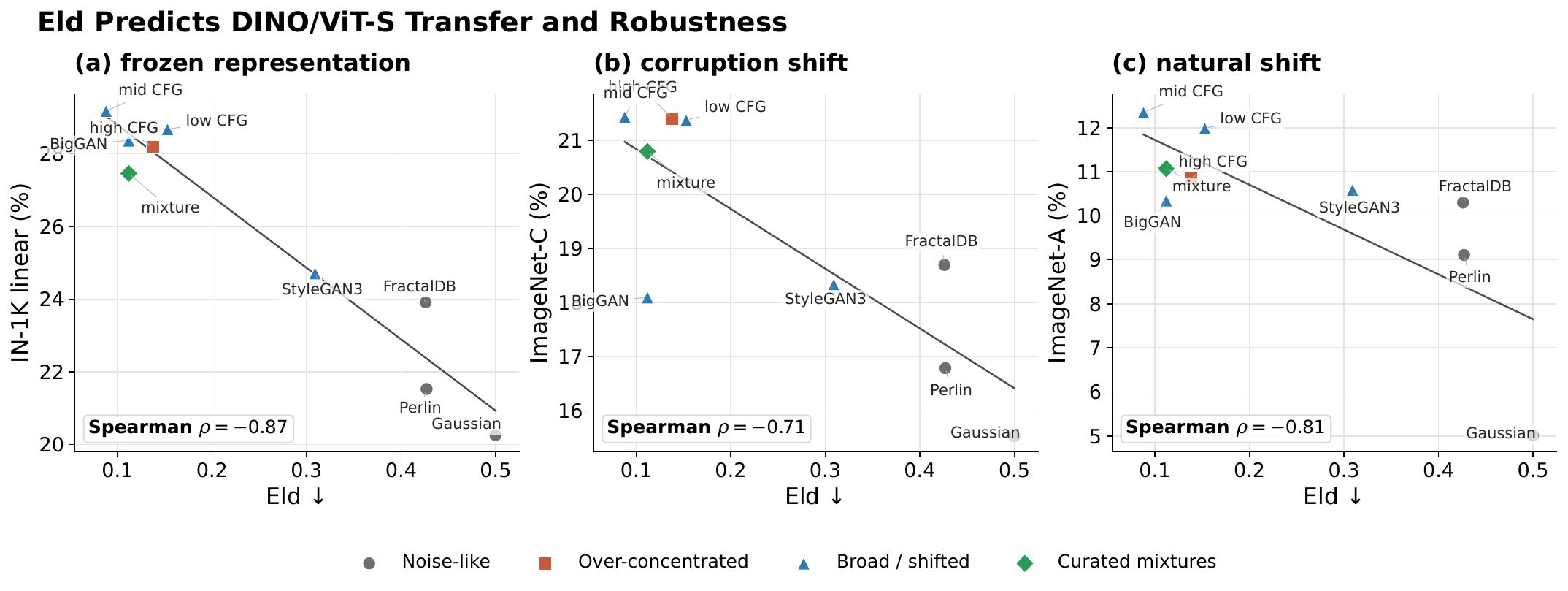}
\caption{\textbf{\textsc{Eld} predicts DINO/ViT-S downstream quality.}
Across the self-supervised synthetic pretraining suite, lower \textsc{Eld} is associated with stronger frozen ImageNet-1K linear-probe accuracy and better robustness on ImageNet-C and ImageNet-A. Points are colored by the energy-regime taxonomy, showing that high-\textsc{Eld} noise-like sources occupy the weak-transfer end of the trend, while lower-\textsc{Eld} diffusion and mixture sources cluster among the strongest representations.}
\vspace{-5pt}
\label{fig:eld-residual}
\end{figure}\label{sec:eld}

\section{Energy Shape Predicts Pretraining Utility}
Sections~\ref{sec:fid-blind} and \ref{sec:eld} argue that energy structure is an axis FID does not see and that \textsc{Eld} measures it. We now ask whether this axis matters: across a broad suite and self-supervised recipes, \textsc{Eld} predicts downstream accuracy beyond FID, LPIPS variation, and recall.

\subsection{Experimental Setup}
\label{sec:exp-setup}
We ask two questions: does \textsc{Eld} predict how well a synthetic dataset trains visual representations, and how does its held-out predictive accuracy compare with that of FID, LPIPS variation, and recall? We evaluate this in two complementary pretraining settings. The first is a self-supervised setting using DINO-style ViT pretraining \citep{caron2021emerging}. The second follows the setting of \citet{baradad2021learning}, where synthetic datasets are used directly as pretraining sources under the original CNN/contrastive evaluation protocol. In both settings, we score each synthetic dataset under the same metric suite and regress downstream accuracy on these metrics. Appendix~\ref{app:synthetic-visualizations} visualizes samples from the corresponding DINO/ViT-S and Learning-to-See synthetic sources.

\textbf{Synthetic pretraining data.} The suite spans the regimes of Section~\ref{sec:synthetic-regimes}: procedural and noise-like sources from \citet{baradad2021learning}, including dead leaves, statistical image models, feature visualizations, and untrained StyleGAN variants; classical synthetic pretraining sources such as FractalDB, Perlin noise, and Gaussian noise; generator-based sources including StyleGAN3, BigGAN, and diffusion/latent-diffusion samples at varied guidance or sampling settings; and curated mixtures such as StableRep and uniform mixtures \citep{baradad2021learning,kataoka2022replacing,karras2021alias,brock2018large,rombach2022high,NEURIPS2023_971f1e59}.

\textbf{Pretraining recipes.}
Our first recipe is DINO with ViT-S under fixed hyperparameters across datasets \citep{caron2021emerging}. 
To connect with earlier synthetic-data pretraining work, we also run the ``Learning to See by Looking at Noise'' protocol using its corresponding CNN-based recipe and synthetic dataset families \citep{baradad2021learning}.

\textbf{Evaluation.} For the self-supervised setting, we evaluate ImageNet-1K representation quality using SimCLR-style linear probing and fine-tuning \citep{pmlr-v119-chen20j}. For the Learning-to-See setting, we follow the original ImageNet100 evaluation protocol of \citet{baradad2021learning}. In both settings, we additionally evaluate distribution shift on ImageNet-C and ImageNet-A \citep{hendrycks2018benchmarking,hendrycks2021natural}; for Learning-to-See, these shift evaluations are restricted to the ImageNet100 class subset and resized to the encoder input resolution. For each pretraining dataset we compute \textsc{Eld} against ImageNet alongside FID, LPIPS, and Improved Recall \citep{NIPS2017_8a1d6947,zhang2018unreasonable,kynkaanniemi2019improved}. Table~\ref{tab:metric-landscape-dino} reports DINO/ViT-S, while Table~\ref{tab:metric-landscape-l2s} reports the Learning-to-See setting. We use DINO/ViT-S as the modern self-supervised protocol and Learning-to-See as a complementary small-scale replication with classical synthetic pretraining.
\begin{table}[t!]
\centering
\small
\caption{\textbf{Self-supervised setting.} Dataset metrics and DINO/ViT-S accuracy; metrics are computed against ImageNet, and evaluation reports ImageNet-1K linear/FT plus ImageNet-C/A robustness.}
\setlength{\tabcolsep}{3pt}
\renewcommand{\arraystretch}{1.05}
\begin{tabular}{lcccccccc}
\toprule
& \multicolumn{4}{c}{\textbf{Data metrics}} & \multicolumn{4}{c}{\textbf{DINO/ViT-S downstream}} \\
\cmidrule(lr){2-5}\cmidrule(lr){6-9}
Dataset & \textsc{\textbf{Eld}}\,$\downarrow$ & FID\,$\downarrow$ & LPIPS\,$\uparrow$ & Recall\,$\uparrow$
& IN-1K Lin.\%\,$\uparrow$ & IN-1K FT\%\,$\uparrow$ & IN-C\%\,$\uparrow$ & IN-A\%\,$\uparrow$ \\
\midrule
\multicolumn{9}{l}{\textit{Noise-like}} \\
FractalDB           & \textbf{.426} & 1254 & .260 & .018 & \textbf{23.91} & \textbf{56.66} & \textbf{18.70} & \textbf{10.30} \\
Perlin noise        & .427 & 1922 & .208 & .000 & 21.53 & 55.38 & 16.79 & 9.11 \\
Gaussian noise      & .500 & 2145 & .237 & .000 & 20.26 & 52.26 & 15.53 & 5.01 \\
\midrule
\multicolumn{9}{l}{\textit{Over-concentrated}} \\
diffusion (high CFG)  & \textbf{.138} & 481 & .342 & .020 & \textbf{28.18} & \textbf{59.34} & \textbf{21.41} & \textbf{10.84} \\
\midrule
\multicolumn{9}{l}{\textit{Broad / shifted}} \\
StyleGAN3           & .309 & 1148 & .305 & .001 & 24.71 & 57.49 & 18.34 & 10.59 \\
BigGAN              & .112 & 566 & .303 & .101 & 28.36 & 58.97 & 18.10 & 10.35 \\
diffusion (mid CFG) & \textbf{.088} & 459 & .338 & .030 & \textbf{29.17} & 59.57 & \textbf{21.44} & \textbf{12.35} \\
diffusion (low CFG) & .153 & 1129 & .297 & .022 & 28.67 & \textbf{59.60} & 21.38 & 11.99 \\
\midrule
\multicolumn{9}{l}{\textit{Curated mixtures}} \\
uniform mixture     & \textbf{.112} & 616 & .294 & .023 & \textbf{27.45} & \textbf{57.61} & \textbf{20.80} & \textbf{11.07} \\
\bottomrule
\end{tabular}
\label{tab:metric-landscape-dino}
\vspace{-10pt}
\end{table}

\subsection{\textsc{Eld} Predicts Downstream Quality Beyond FID}
\label{sec:exp-correlation}
We next test whether probability-level mismatch predicts pretraining utility in a modern self-supervised setting. Table~\ref{tab:metric-landscape-dino} reports DINO/ViT-S models pretrained on each synthetic source, with ImageNet-1K linear probing, ImageNet-1K fine-tuning, and ImageNet-C/A robustness. Across this suite, \textsc{Eld} is consistently negatively associated with downstream performance: lower \textsc{Eld} corresponds to higher frozen ImageNet-1K linear accuracy (Spearman $\rho=-0.87$, $p=0.002$), higher ImageNet-1K fine-tuning accuracy ($\rho=-0.78$, $p=0.014$), and better robustness on ImageNet-C ($\rho=-0.71$, $p=0.032$) and ImageNet-A ($\rho=-0.81$, $p=0.008$). These results support the directional prediction of Section~\ref{sec:synthetic-regimes}: sources whose energy profiles depart from the real-image likelihood distribution yield weaker transferable representations. Figure~\ref{fig:eld-residual} visualises the trend, with regimes organizing the scatter from high-\textsc{Eld} noise-like sources at the weak-transfer end to lower-\textsc{Eld} diffusion and mixture sources among the strongest models.

\paragraph{Held-out prediction.} FID and recall also correlate with transfer on the DINO suite. To compare predictive utility, we fit a separate univariate linear regression for each metric, omitting one dataset at a time and predicting its downstream accuracy. ELD gives the lowest prediction error for DINO linear accuracy among the four evaluated metrics (Table~\ref{tab:dino_prediction}). This advantage persists after excluding Gaussian noise, Perlin noise, and FractalDB: ELD's mean absolute error is 0.99 percentage points, compared with 1.66 for FID, 1.47 for LPIPS variation, and 3.27 for recall. Full comparisons and analyses appear in Appendix~\ref{app:prediction}.
\begin{table}[h!]
\centering
\setlength{\tabcolsep}{8pt}
\renewcommand{\arraystretch}{1.05}
\vspace{-12pt}
\caption{Leave-one-dataset-out DINO prediction MAE (percentage points $\downarrow$).}
\begin{tabular}{lcccc}
\toprule
DINO linear evaluation & ELD & FID & LPIPS & Recall \\
\midrule
All sources ($n=9$) & \textbf{0.82} & 0.91 & 1.67 & 4.53 \\
Without noise-like sources ($n=6$)
 & \textbf{0.99} & 1.66 & 1.47 & 3.27 \\
\bottomrule
\end{tabular}
\label{tab:dino_prediction}
\vspace{-8pt}
\end{table}

\begin{table}[h!]
\centering
\caption{\textbf{Learning-to-See synthetic-pretraining setting.} Dataset metrics and downstream accuracy for the small-scale families from \citet{baradad2021learning}. Evaluation follows the original ImageNet protocol and adds ImageNet-C/A robustness with the same representations.}
\small
\setlength{\tabcolsep}{3pt}
\renewcommand{\arraystretch}{1.05}
\begin{tabular}{lcccccccc}
\toprule
& \multicolumn{4}{c}{\textbf{Data metrics}} & \multicolumn{4}{c}{\textbf{Learning-to-See downstream}} \\
\cmidrule(lr){2-5}\cmidrule(lr){6-9}
Dataset & \textsc{\textbf{Eld}}\,$\downarrow$ & FID\,$\downarrow$ & LPIPS\,$\uparrow$ & Recall\,$\uparrow$
& IN Lin.\%\,$\uparrow$ & IN FT\%\,$\uparrow$ & IN-C\%\,$\uparrow$ & IN-A\%\,$\uparrow$ \\
\midrule
\multicolumn{9}{l}{\textit{Dead leaves}} \\
dead-leaves squares      & .393 & 1480 & .316 & .005 & 38.60 & 67.52 & 27.23 & 3.24 \\
dead-leaves oriented     & .319 & 1390 & .299 & .035 & 40.90 & 68.48 & 25.12 & \textbf{3.97} \\
dead-leaves mixed        & .310 & 1310 & .301 & .074 & 40.84 & \textbf{68.86} & 25.00 & 3.82 \\
dead-leaves textures     & \textbf{.261} & 1340 & .241 & .084 & \textbf{41.00} & 68.20 & \textbf{27.93} & 3.83 \\
\midrule
\multicolumn{9}{l}{\textit{Statistical image models}} \\
spectrum                 & .314 & 1560 & .159 & .003 & 30.70 & 62.20 & 24.57 & 3.68 \\
WMM                      & .172 & 1520 & .152 & .003 & 34.08 & 63.30 & 16.90 & 3.53 \\
spectrum+color           & .251 & 1290 & .258 & .001 & 37.88 & 65.18 & 27.82 & 4.12 \\
spectrum+color+WMM       & \textbf{.049} & 1210 & .228 & .010 & \textbf{38.84} & \textbf{65.62} & \textbf{30.28} & \textbf{4.56} \\
\midrule
\multicolumn{9}{l}{\textit{Untrained StyleGAN}} \\
StyleGAN random          & .479 & 1420 & .279 & .000 & 35.30 & 64.82 & 29.10 & \textbf{4.41} \\
StyleGAN high-freq.      & \textbf{.085} & 1320 & .268 & .002 & \textbf{39.62} & \textbf{66.21} & \textbf{29.84} & 4.38 \\
StyleGAN sparse          & .278 & 1240 & .305 & .001 & 38.76 & 66.16 & 27.23 & 3.24 \\
StyleGAN oriented        & .355 & 903 & .272 & .007 & 39.08 & 65.84 & 27.70 & 3.38 \\
\midrule
\multicolumn{9}{l}{\textit{Feature visualizations}} \\
feature-vis random       & .402 & 1590 & .252 & .000 & 35.26 & 66.22 & 22.89 & 2.65 \\
feature-vis dead leaves  & \textbf{.365} & 1400 & .223 & .001 & \textbf{38.70} & \textbf{68.00} & \textbf{32.04} & \textbf{4.71} \\
\bottomrule
\end{tabular}
\vspace{5pt}
\label{tab:metric-landscape-l2s}
\vspace{-10pt}
\end{table}

\begin{figure}[t]
\centering
\includegraphics[width=0.95\linewidth]{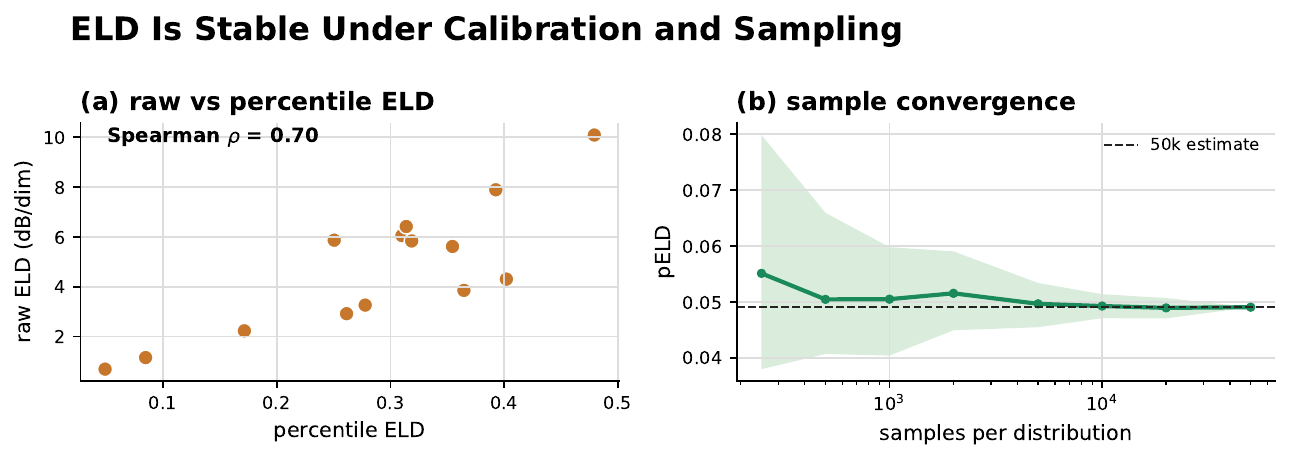}
\caption{\textbf{\textsc{Eld} is robust to estimation choices.}
Raw and percentile distances agree with a Spearman coefficient of $0.7$; subsampling shows stable estimates after a few thousand scored samples.}
\vspace{-15pt}
\label{fig:robustness}
\end{figure}

We also evaluate \textsc{Eld} in the classical synthetic-pretraining setting of \citet{baradad2021learning}. Table~\ref{tab:metric-landscape-l2s} reports the small-scale synthetic families under the original ImageNet100 transfer protocol, augmented with ImageNet-C/A evaluations. This benchmark is heterogeneous by construction: dead leaves, statistical models, untrained StyleGANs, and feature visualizations differ not only in energy profile but also in construction mechanism and semantics. Comparing across heterogeneous synthetic sources, the pooled correlations remain directionally consistent: lower \textsc{Eld} is associated with higher ImageNet-A accuracy (Spearman $\rho=-0.26$) and weakly with higher ImageNet-C accuracy ($\rho=-0.15$). ImageNet-A accuracy is computed on the intersection with the trained ImageNet100 classes; examples outside this label set are excluded from the evaluation denominator.

The within-family comparisons are particularly informative. Among dead-leaves variants, lower \textsc{Eld} tracks stronger early ImageNet transfer ($\rho=-0.80$ for linear accuracy), with the lowest-\textsc{Eld} texture variant achieving the best linear accuracy and ImageNet-C accuracy in the group. Among statistical image models, \textsc{Eld} ranks both ImageNet linear and fine-tuned transfer strongly ($\rho=-0.80$ for each), and the best energy match, spectrum+color+WMM, is also the strongest source on ImageNet linear transfer, fine-tuning, ImageNet-C, and ImageNet-A. The untrained StyleGAN family shows the same pattern for ImageNet transfer: lower \textsc{Eld} is associated with stronger linear accuracy ($\rho=-0.80$) and fine-tuned accuracy ($\rho=-1.00$), with the high-frequency variant attaining the best transfer scores. For the two feature-visualization sources, the lower-\textsc{Eld} dead-leaves variant is better on every downstream metric. \textsc{Eld} provides a consistent directional signal especially when comparing synthetic sources within a shared family.

Together, the two experimental settings support the regime picture of Section~\ref{sec:synthetic-regimes}. \textsc{Eld} does not subsume all factors governing synthetic pretraining: semantic coverage, augmentation compatibility, and feature diversity remain important. Instead, it isolates a complementary axis of mismatch, namely whether a synthetic source places probability mass at the likelihood levels occupied by real images. The DINO/ViT-S results show that this axis predicts transfer and robustness in a modern self-supervised protocol, while the Learning-to-See results show that the same signal remains interpretable in a classical small-scale synthetic-pretraining setting, especially within matched generator families. This is the role of \textsc{Eld} in our analysis: not a universal replacement for feature-space metrics, but a probability-level diagnostic that explains a structured component of downstream variation.

\paragraph{Estimation robustness.} Figure~\ref{fig:robustness} verifies that the reported percentile version is not an artefact of calibration or sample size. Raw energy-space distance and percentile \textsc{Eld} are strongly aligned across datasets, indicating that rank calibration preserves the relevant ordering while removing dependence on the absolute scale of the density model. Subsampling shows that \textsc{Eld} estimates stabilise after a few thousand scored examples, well below the sample sizes used in the main experiments.

\paragraph{Robustness to the neural scorer.} We recompute pELD for the nine DINO sources with two additional neural scoring constructions, calibrating each against its own ImageNet reference. Dataset rankings closely agree with those of the primary Guth scorer (Spearman $\rho=0.89$--$0.91$), and lower pELD remains associated with stronger transfer (Table~\ref{tab:neural_scorers}). This supports the stability of the observed signal across the tested neural scorers.

\begin{table}[t]
\centering
\caption{Neural-scorer sensitivity on the DINO suite. The first
column compares dataset rankings; the remaining columns correlate
pELD with ImageNet-1K accuracy.}
\begin{tabular}{lccc}
\toprule
Scorer & Agreement with Guth $\rho$ & Linear $\rho$ & Fine-tuning $\rho$ \\
\midrule
Guth dual-score model & 1.00 & -0.87 & -0.78 \\
DenseFlow & 0.89 & -0.85 & -0.78 \\
Diffusion VLB & 0.91 & -0.77 & -0.72 \\
\bottomrule
\end{tabular}
\label{tab:neural_scorers}
\vspace{-10pt}
\end{table}

\label{sec:experiments}

\section{Energy-Aware Curation for Synthetic Data}
\textsc{Eld} suggests a constructive use beyond evaluation: synthetic data can be \emph{shaped} to match the probability-level profile of real images. Natural images occupy a broad energy spectrum containing both simple high-likelihood scenes and dense low-likelihood structure \citep{guth2026learning,field1987relations,ruderman1997origins,simoncelli20054}. Filtering only for the highest-likelihood samples can therefore discard variation present in real data. The goal is to make synthetic pretraining data populate real-image likelihood levels in the right proportions, not only to retain the most probable samples.

Because real-energy percentiles are uniform under a continuous reference CDF, a simple curation rule is to resample synthetic data so that its percentile histogram is approximately uniform. In practice, we can score a candidate pool under $p_P$, bin samples by their real-energy percentiles, and draw a balanced training stream across bins, with complementary diversity criteria guiding selection within each band. This changes the composition of the selected set without retraining a generator. Reweighting cannot recover energy bands absent from the candidate pool; such gaps require additional sources or generation settings. For mixtures, the same principle becomes a small convex design problem: choose source weights whose combined energy histogram best matches the real one under a convex histogram discrepancy, such as discretised Wasserstein-1 distance. Thus the audit becomes prescriptive: over-concentrated sources can be supplemented with underrepresented higher-energy structure; noise-like sources can be downweighted in favour of candidates in reference-supported energy bands; and broad but shifted sources can contribute where they fill missing likelihood bands. These interventions target energy occupancy, while semantic and feature-space diagnostics remain necessary to assess the resulting composition.

\paragraph{Controlled offline curation.} Energy-profile matching is an alternative to selecting only the highest-likelihood images. We compare source-balanced uniform selection, top-likelihood filtering, and ELD-guided selection from a common pool spanning BigGAN, StyleGAN3, and three diffusion settings, using the same selected-set budget across strategies. Top-likelihood filtering retains the lowest-energy candidates; ELD-guided selection seeks to match the full ImageNet energy-percentile histogram. No generator retraining or downstream model fitting is required for this offline intervention.

We evaluate the selected sets using frozen ImageNet-pretrained ResNet features (Table~\ref{tab:offline_curation}). ELD-guided selection improves both reported nearest-neighbour distance diagnostics and increases effective class coverage and feature rank relative to uniform selection. Top-likelihood filtering performs worse on all four diagnostics. Thus matching energy occupancy can retain useful feature diversity that likelihood maximisation discards. These are offline proxy results; matched pretraining experiments are deferred to future work to further establish a transfer gain caused by the intervention.

\begin{table}[t] 
\centering  
\caption{Controlled offline curation from a common source pool at equal selected-set budget. Frozen-feature diagnostics favour ELD-guided selection over uniform and top-likelihood selection. NN denotes the cosine-distance nearest-neighbour diagnostic.}
\begin{tabular}{lcccc} 
\toprule 
Selection & NN mean $\downarrow$ & NN p95 $\downarrow$ & Eff. classes $\uparrow$ & Eff. rank $\uparrow$ \\ 
\midrule 
Uniform & 0.634 & 0.784 & 116.445 & 59.175 \\ 
Top likelihood & 0.660 & 0.788 & 82.940 & 34.132 \\ 
ELD-guided & \textbf{0.621} & \textbf{0.780} & \textbf{128.001} & \textbf{71.828} \\ 
\bottomrule 
\end{tabular}
\vspace{-8pt}
\label{tab:offline_curation} 
\end{table}

More broadly, \textsc{Eld} reframes synthetic-data construction as distributional engineering. Given a synthetic pool with cached energies and features, we may select samples to match the real energy histogram while imposing complementary feature-coverage or diversity constraints within each band. The selected distribution $Q_{\mathrm{sel}}$ should satisfy \[ (E_P)_\#Q_{\mathrm{sel}}\approx(E_P)_\#P, \] while retaining useful variation along dimensions not determined by energy alone. This seeks to restore under-covered likelihood bands, rather than merely keep the most likely samples. It turns common design choices---which samples to retain, which generators to mix, and which prompts or guidance scales to use---into ways of redistributing mass along the real-image energy axis. The target along this axis is the full probability-level profile of natural images; energy matching complements, rather than replaces, semantic and task-specific selection criteria.\label{sec:curation}

\section{Conclusion}
\paragraph{Discussion and Limitations.} \textsc{Eld} makes synthetic data inspectable at the level of likelihood structure, revealing which parts of a reference probability landscape samples occupy. It complements FID, recall, and LPIPS variation: feature moments can miss energy-profile differences, while energy matching alone does not guarantee semantic coverage, diversity, label quality, or task relevance. Combining these criteria supports audits of both likelihood structure and feature-space realism and coverage. The reference data and density model are part of the experimental specification, so \textsc{Eld} should be reported with its scorer. Percentile calibration removes strictly increasing score recalibrations; our cross-scorer analyses support robustness across the tested neural scoring constructions. Our experiments focus on visual pretraining; supervised-label efficiency, aesthetic fidelity, and human preference may require additional task-aware criteria. CLIP-style multimodal pretraining further motivates joint audits of visual, linguistic, and image--text alignment structure.

Synthetic data is not characterised by visual realism or feature-space similarity alone; its position in the real-image probability landscape also matters. \textsc{Eld} measures this axis by comparing the distribution of likelihood levels induced by real and synthetic data under a shared density model. We show that synthetic sources occupy distinct energy regimes that feature moments do not determine, and that these regimes are predictive of downstream pretraining quality in the studied settings. Our controlled curation study demonstrates a practical intervention through improved frozen-feature diagnostics; matched pretraining runs can test how these gains translate into transfer. Since the mismatch is structured, \textsc{Eld} is also prescriptive: it identifies likelihood bands that are missing, overrepresented, or shifted, enabling resampling and mixing strategies that better match the reference energy profile. More broadly, this supports a design view of synthetic data: generators, filters, and mixtures can be selected for how they populate the real-image likelihood spectrum, alongside their realism, diversity, and task relevance.

\section*{Acknowledgments}
This work was partially completed during C.L.'s visit to the Technical University of Munich, supported by the Munich Center for Machine Learning (MCML). S.J. was supported by an Alexander von Humboldt professorship. A.T. was supported by the ONR MURI grant N000142612255. This work was further supported by a Packard Fellowship to P.I., along with the ONR MURI grant N00014-22-1-2740. 

\bibliography{ref}
\bibliographystyle{ref}

\newpage
\appendix
\onecolumn
\section{Feature Moments and Energy Profiles}
\label{app:proofs}
\label{app:fid_fibres}

\paragraph{Proposition.}
Let $\phi:\mathcal X\to\mathbb R^d$ be a measurable feature map, and let $E_P:\mathcal X\to\mathbb R$ be a fixed reference energy. Let $\nu$ be a feature distribution with finite second moments. Suppose there exist measurable selections $x^-,x^+:\mathbb R^d\to\mathcal X$ such that
\[
    \phi(x^-(z))=\phi(x^+(z))=z
    \qquad \text{for $\nu$-almost every }z.
\]
Define $Q^\pm=(x^\pm)_\#\nu$. Then $Q^-$ and $Q^+$ have identical feature distributions, and hence identical FID to any fixed reference with finite feature second moments. Their energy profiles nevertheless differ whenever
\[
    (E_P\circ x^-)_\#\nu
    \neq
    (E_P\circ x^+)_\#\nu.
\]

Moreover, suppose the induced energies have finite first moments. If a measurable feature set $Z$ has mass $\nu(Z)=\alpha$, the selections agree outside $Z$, and
\[
    E_P(x^+(z))-E_P(x^-(z))\geq\Delta>0
    \qquad \text{for $\nu$-almost every }z\in Z,
\]
then
\[
    W_1\!\left((E_P)_\#Q^-,(E_P)_\#Q^+\right)
    \geq \alpha\Delta.
\]

\paragraph{Proof.}
By construction, $\phi_\#Q^-=\phi_\#Q^+=\nu$. Thus both distributions have the same feature mean and covariance, yielding identical FID to any fixed reference. Their energy pushforwards are $(E_P)_\#Q^\pm=(E_P\circ x^\pm)_\#\nu$, which differ under the stated hypothesis.

For the quantitative claim, agreement outside $Z$ gives
\[
\begin{aligned}
    \mathbb E_{Q^+}[E_P]-\mathbb E_{Q^-}[E_P]
    &=
    \int_Z
    \bigl(E_P(x^+(z))-E_P(x^-(z))\bigr)\,d\nu(z)\\
    &\geq \alpha\Delta.
\end{aligned}
\]
The Wasserstein-1 distance is at least the absolute difference of the means, proving the bound.
\hfill$\square$

The proposition establishes that feature moments do not determine energy occupancy. Conversely, matching an energy profile does not determine the image distribution or guarantee semantic coverage; the two diagnostics therefore provide complementary information.

\section{Experiments}
\paragraph{Compute details.} All density scoring, metric computation, and downstream training jobs were run on NVIDIA V100 GPUs. Energy scores are computed once per dataset and reused across all \textsc{Eld} analyses; after scoring, \textsc{Eld} itself is computed from sorted one-dimensional energy values and is negligible compared with image scoring or representation training.
\begin{table}[h!]
\centering
\vspace{-5pt}
\caption{\textbf{Coarse regime diagnostics.} Mean shift and spread separate the main regimes; \textsc{pEld} captures distributional mismatch beyond these summaries.}
\setlength{\tabcolsep}{7pt}
\renewcommand{\arraystretch}{1.08}
\begin{tabular}{lcccc}
\toprule
Regime & \# datasets & Mean shift & Mean spread & Mean \textsc{pEld} \\
\midrule
Over-concentrated        & 3 & -4.43 & 0.52 & .281 \\
Broad / shifted          & 7 & -5.41 & 1.47 & .284 \\
Noise-like / high-energy & 4 &  2.92 & 0.22 & .300 \\
\bottomrule
\end{tabular}
\label{tab:appendix-regime-summary}
\vspace{-15pt}
\end{table}

\subsection{Diagnostics}
\paragraph{Simple diagnostics give a coarse regime map, not a substitute for \textsc{Eld}.} Mean energy shift and spread ratio provide a useful low-dimensional summary of how a source departs from ImageNet: they separate compression, displacement, and high-energy mismatch. Table~\ref{tab:appendix-regime-summary} shows that these summaries recover the coarse regimes used in the main text. However, they do not determine the full energy distribution. Sources within the same regime can have different tail structure and different \textsc{pEld}, so the Wasserstein comparison remains necessary for quantifying residual distributional mismatch.

\begin{table}[h]
\centering
\caption{\textbf{Simple surrogate energies do not reproduce \textsc{Eld}.} Feature- and DCT-based density surrogates fail to recover the ranking induced by the learned ImageNet density model, indicating that \textsc{Eld} is not reducible to generic feature novelty or low-level image-statistics mismatch.}
\small
\setlength{\tabcolsep}{3pt}
\renewcommand{\arraystretch}{1.08}
\begin{tabular}{lcc}
\toprule
Surrogate energy model & Description & Spearman $\rho$ with \textsc{pELD} \\
\midrule
ResNet-feature Gaussian & Diagonal Gaussian on frozen features & $-0.22$ \\
DCT Gaussian & Diagonal Gaussian on low-frequency DCT coefficients & $-0.28$ \\
DCT-GMM & GMM on PCA-compressed DCT coefficients & $-0.26$ \\
\bottomrule
\end{tabular}
\label{tab:surrogate-energy-ablation}
\vspace{-5pt}
\end{table}

\paragraph{Generic feature-density energies are not substitutes for \textsc{Eld}.} We also test whether simpler surrogate energy models reproduce the ranking induced by the learned ImageNet density. Specifically, we fit three lightweight alternatives on ImageNet: a diagonal Gaussian over frozen ResNet-18 features, a diagonal Gaussian over low-frequency greyscale DCT coefficients, and a Gaussian mixture model over PCA-compressed DCT features. Table~\ref{tab:surrogate-energy-ablation} reports their rank agreement with the primary scorer on the small-scale synthetic suite. These simple surrogates do not recover its dataset ordering, supporting the use of the learned real-image density scorer for this audit. Their rank correlations with primary \textsc{pEld} are negative or near zero: $\rho=-0.22$ for ResNet-feature Gaussian energy, $\rho=-0.28$ for diagonal DCT-Gaussian energy, and $\rho=-0.26$ for DCT-GMM energy. This suggests that \textsc{Eld} is not merely measuring generic feature outlierness, spectrum mismatch, or distance from simple image statistics. The choice of $p_P$ matters: \textsc{Eld} is intended to compare datasets under a real-image density model, not under arbitrary feature-space novelty scores.

\subsection{Synthetic Source Visualizations}
\label{app:synthetic-visualizations}
\begin{figure}[t]
\centering
\includegraphics[width=\linewidth]{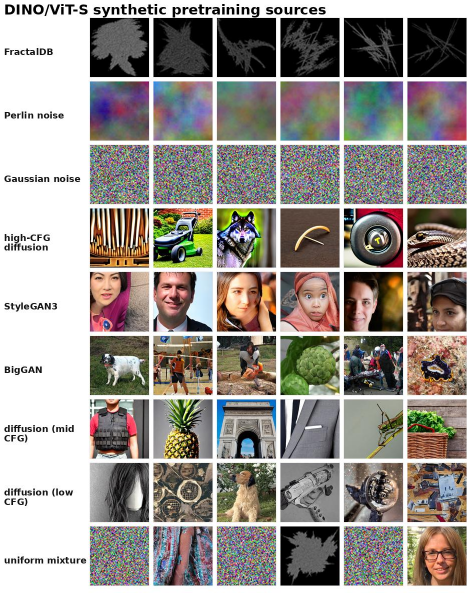}
\caption{\textbf{DINO/ViT-S synthetic pretraining sources.}
This grid highlights the range from noise and procedural structure to generator samples; \textsc{Eld} measures how these sources populate ImageNet likelihood levels.}
\vspace{-5pt}
\label{fig:appendix-synthetic-samplesA}
\end{figure}

\begin{figure}[t]
\centering
\includegraphics[width=\linewidth]{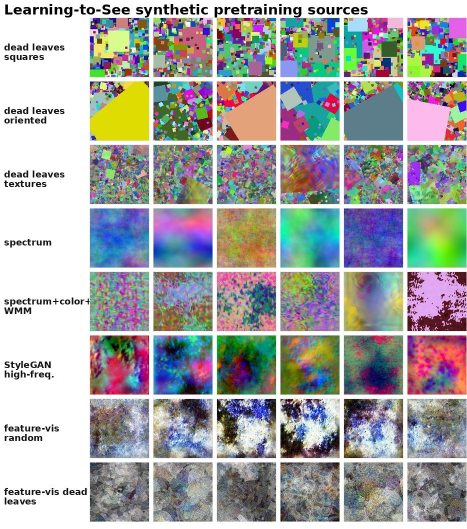}
\caption{\textbf{Learning-to-See synthetic pretraining sources.} This grid visualises the range from dead-leaves and statistical textures to untrained StyleGAN and feature-visualization samples; \textsc{Eld} measures how these families populate ImageNet likelihood levels.}
\vspace{-5pt}
\label{fig:appendix-synthetic-samplesB}
\end{figure}

Figures~\ref{fig:appendix-synthetic-samplesA} and~\ref{fig:appendix-synthetic-samplesB} show sampled images from the synthetic sources used in the DINO/ViT-S and Learning-to-See experiments. The visual differences across sources are substantial, but visual appearance alone does not determine the energy profile: visually diverse generators can still be shifted or compressed along the real-image likelihood axis, while simpler sources can occupy distinct energy regimes. These grids provide qualitative context for the quantitative energy audit in the main text.

\begin{table}[h!]
\centering
\caption{Sensitivity of DINO linear prediction to source exclusion.
Each row repeats leave-one-dataset-out fitting within the retained
subset. It does not evaluate transfer to the omitted family.}
\setlength{\tabcolsep}{8pt}
\renewcommand{\arraystretch}{1.15}
\begin{tabular}{lccccc}
\toprule
DINO subset & $n$ & ELD & FID & LPIPS & Recall \\
\midrule
All sources & 9 & \textbf{0.82} & 0.91 & 1.67 & 4.53 \\
Without noise-like & 6 & \textbf{0.99} & 1.66 & 1.47 & 3.27 \\
Without GANs & 7 & \textbf{1.06} & 1.17 & 1.63 & 1.34 \\
Without diffusion & 6 & 1.07 & \textbf{0.46} & 2.26 & 4.08 \\
Without uniform mixture & 8 & \textbf{0.76} & 0.96 & 1.75 & 4.95 \\
\bottomrule
\end{tabular}
\label{tab:restricted_prediction}
\vspace{-5pt}
\end{table}

\subsection{Held-Out Prediction and Source-Exclusion Analyses}
\label{app:prediction}
We examine whether the DINO linear-accuracy prediction result depends on particular synthetic sources. Table~\ref{tab:restricted_prediction} compares ELD, FID, LPIPS variation, and recall on the full suite and after excluding noise-like sources, GANs, diffusion sources, or the uniform mixture. Within each retained subset, we repeat leave-one-dataset-out prediction using a separate ordinary least-squares regression with an intercept for each metric, with no hyperparameter tuning. Each evaluated dataset is excluded from regression fitting. MAE is measured in accuracy percentage points. These comparisons assess sensitivity to source composition, not training-seed variability or generalisation to an omitted family.

ELD has the lowest MAE on the full suite and after excluding noise-like sources, GANs, or the uniform mixture. In particular, its advantage after removing noise-like sources supports a predictive signal beyond the separation of obvious low-quality datasets. The comparison remains family-dependent: FID performs better after diffusion sources are removed. These small-suite analyses support energy occupancy as a useful predictor of DINO linear transfer across several source compositions, alongside its diagnostic interpretation.

\end{document}